\documentclass{article}

\usepackage[eandd,preprint]{neurips_2026}

\usepackage[utf8]{inputenc}
\usepackage[T1]{fontenc}
\usepackage{hyperref}
\usepackage{url}
\usepackage{booktabs}
\usepackage{amsfonts}
\usepackage{nicefrac}
\usepackage{microtype}
\usepackage{xcolor}
\usepackage{graphicx}
\usepackage{tabularx}
\usepackage{amsmath}

\newcommand{\dataset}{Physical AI Smart Spaces}
\newcommand{\paiss}{PAISS}

\title{\dataset: A Large-Scale Benchmark for Multi-Camera 3D Perception in Smart Spaces}

\author{
\normalfont\mdseries
Yuxing Wang\textsuperscript{1},
Yizhou Wang\textsuperscript{1},
Anqi Li\textsuperscript{1},
Shuo Wang\textsuperscript{1},
Sameer Satish Pusegaonkar\textsuperscript{1},
\\
Haoquan Liang\textsuperscript{1},
Jiajun Li\textsuperscript{1},
Shenxin Jiang\textsuperscript{1},
Jianhe Yuan\textsuperscript{1},
Shangru Li\textsuperscript{1},
Tongwei Dai\textsuperscript{1},
\\
Zihao Chen\textsuperscript{1},
David C. Anastasiu\textsuperscript{2},
Sujit Biswas\textsuperscript{1},
Xunlei Wu\textsuperscript{1},
Zheng Tang\textsuperscript{1,*}
\\[0.75em]
\textsuperscript{1}NVIDIA, Santa Clara, CA, USA
\\
\textsuperscript{2}Santa Clara University, Santa Clara, CA, USA
\\[0.4em]
\textsuperscript{*}Corresponding author:
Zheng Tang, \texttt{thtang@nvidia.com}
}

\begin{document}

\maketitle

\begin{abstract}
\dataset{} is, to the best of our knowledge, the first benchmark to simultaneously provide large-scale, multi-class, and multi-camera 3D perception data for indoor smart spaces. It contains over 280 hours of synchronized 1080p footage captured by nearly 1,800 cameras in warehouses, hospitals, retail venues, and similar settings, together with automatic annotations for multi-camera identities, 2D bounding boxes, 3D bounding boxes, camera calibration, and depth where available. The benchmark spans Isaac Sim synthetic generation, Cosmos Transfer appearance augmentation, and real-world Sim2Real evaluation. For the real-world target, we include two warehouse deployments with time-synchronized streams, automatic VGGT-based calibration, and a 3D labeling interface that projects world-frame 3D boxes into each view for cross-camera verification. We describe the dataset scope, annotation and calibration schema, generation workflow, benchmark protocols, and official evaluation system, which standardizes submission format, and leaderboard reporting. A central contribution is a 3D instantiation of Higher Order Tracking Accuracy (HOTA), extending the usual 2D box-based tracking evaluation to 3D locations and 3D boxes. We further report empirical baselines from the AI City Challenge leaderboards, showing how methods evolve from person-only 3D location tracking to multi-class 3D box tracking under realistic smart-space constraints. The release is available at \url{https://huggingface.co/datasets/nvidia/PhysicalAI-SmartSpaces}.
\end{abstract}

\section{Introduction}
\label{sec:intro}

Multi-camera perception is a core capability for physical AI systems operating in warehouses, hospitals, retail spaces, factories, and transportation hubs. These environments require systems to detect objects in each view, reason about their 3D locations, and maintain consistent identities as persons, forklifts, pallet trucks, autonomous mobile robots (AMRs), and humanoid platforms move through overlapping and non-overlapping camera networks. Evaluation in this setting is difficult: real deployments are privacy-sensitive, manual multi-view 3D annotation is expensive, and benchmark datasets often trade off scale, calibration quality, object diversity, and reproducible evaluation.

\begin{figure}[t]
  \centering
  \includegraphics[width=\linewidth]{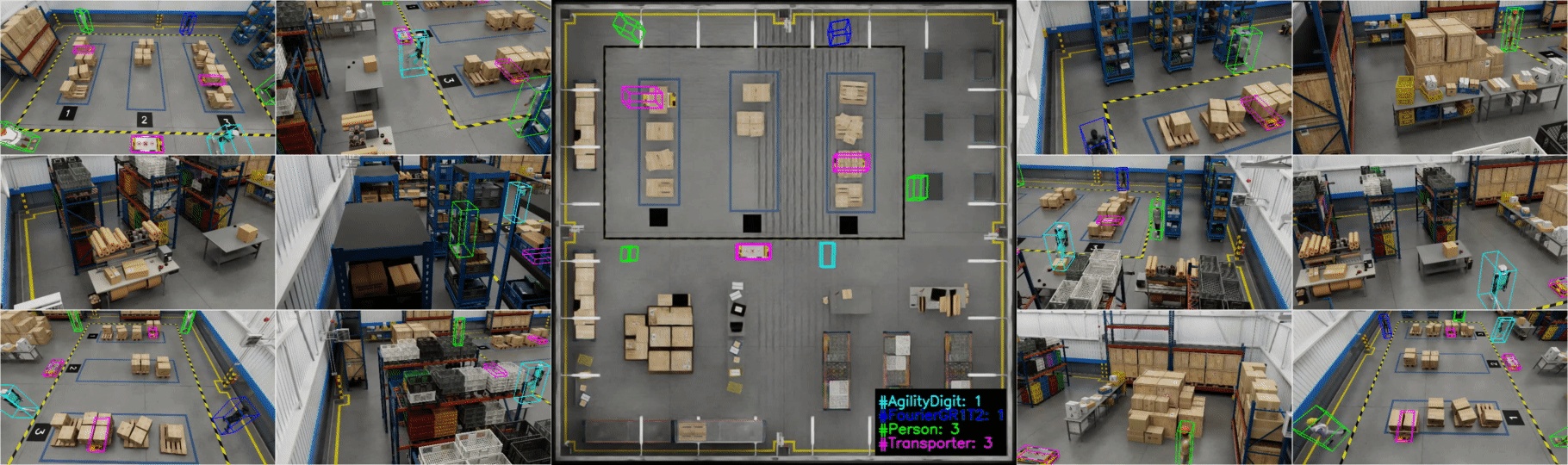}
  \vspace{-6mm}
  \caption{Example synchronized views and top-down scene context from \dataset{} (\paiss). The dataset provides multi-camera RGB streams together with geometry-aware annotations for 3D perception.}
  \label{fig:paiss_demo}
  \vspace{-5mm}
\end{figure}

This paper presents \dataset{} (\paiss) as a large-scale benchmark dataset for multi-class and multi-camera 3D perception. The dataset is publicly hosted as \texttt{nvidia/PhysicalAI-SmartSpaces}~\citep{physicalai_smartspaces_hf} and has served as Track 1 of the AI City Challenge in 2024, 2025, and 2026~\citep{wang20248th,tang20259th,aicity2026track1}. Across these releases, the dataset includes 139 multi-camera scenes, 282.5 hours of synchronized video, 1,799 camera streams, more than 72 million 3D object annotations or locations, and more than 241 million 2D boxes. Figure~\ref{fig:paiss_demo} illustrates the synchronized multi-view structure. The data is generated primarily through NVIDIA Omniverse and Isaac Sim tooling, with dense labels produced automatically rather than through manual annotation~\citep{nvidia_isaacsim_docs}.

The central contribution is not only scale. \dataset{} is intended to support evaluation of a concrete claim: multi-camera perception systems should generalize from controlled synthetic supervision to realistic deployment conditions while preserving 3D localization and identity consistency. The 2026 benchmark makes this evaluative role explicit by adding Cosmos Transfer 2.5 (CT2.5) augmented RGB renderings and a hidden real-world target. The CT2.5 scenes preserve geometry, calibration, depth, and ground truth while changing the RGB appearance, enabling controlled studies of visual domain shift. The hidden real-world target evaluates whether models trained on synthetic and appearance-augmented data remain effective when depth is unavailable at inference time.

The benchmark makes four contributions:
\begin{itemize}
  \item It provides a large-scale public benchmark for synchronized multi-camera 3D perception in indoor smart spaces with multi-class object annotations including persons, AMRs, humanoids, forklifts, etc.
  \item It couples automatic 2D/3D annotation, calibration, depth, and multi-camera identity labels with a documented generation pipeline for reproducible synthetic data creation.
  \item It adds Cosmos-Transferred data augmentation technique and a hidden real-world target to improve Sim2Real generalization under RGB-only inference constraints.
  \item It introduces a 3D HOTA evaluation protocol that evaluates location- and box-based 3D tracking rather than conventional 2D box tracking alone.

\end{itemize}

\section{Related Works}
\label{sec:related}

Multi-target multi-camera (MTMC) tracking datasets have historically emphasized person re-identification, 2D localization, and identity association across sparse camera networks. Early public benchmarks such as PETS2009~\citep{ferryman2009pets}, USC Campus~\citep{javed2008modeling}, the EPFL/Passageway sequences~\citep{berclaz2011multiple}, NLPR\_MCT~\citep{nlprmct2014}, and CamNet~\citep{zhang2015camnet} helped establish cross-camera association as a reproducible research task, but were limited in scale and annotation richness. Later resources such as DukeMTMC~\citep{ristani2016performance}, WILDTRACK~\citep{chavdarova2018wildtrack}, MTA~\citep{kohl2020mta}, MultiviewX~\citep{hou2020multiviewx}, MMPTRACK~\citep{han2023mmptrack}, MTMMC~\citep{woo2024mtmmc}, and SCOUT~\citep{engilberge2025unified} expanded scale, resolution, synthetic coverage, or real-world diversity. These datasets remain important, but they still leave gaps for physical AI deployments: multi-class mobile objects, dense 3D supervision, explicit depth and calibration, and controlled synthetic-to-real evaluation.

Synthetic data has become increasingly important for embodied AI and smart-space perception because it can provide labels that are impractical to collect at scale in physical environments. \dataset{} follows this direction by using simulator-generated synchronized videos and automatic labels, while retaining the challenge-style evaluation discipline of public protocols and hidden test sets. The benchmark also connects to recent world generation and transfer models: Cosmos Transfer style augmentation is used to produce appearance-shifted renderings while preserving geometry and labels~\citep{nvidia2025cosmostransfer1conditionalworldgeneration,nvidia2025worldsimulationvideofoundation}. This gives the dataset an evaluative role beyond training data volume: it enables controlled measurement of whether visual realism and domain transfer improve multi-camera 3D tracking.

Table~\ref{tab:dataset_comparison_smartspaces} compares \paiss{} with representative public multi-camera benchmarks. The table emphasizes the axes that matter for physical AI smart spaces: camera-network scale, overlapping and non-overlapping topology, 3D supervision, depth availability, and high-resolution synchronized video. \paiss{} is larger in camera count and video duration than most prior MTMC resources, but its main distinction is the combination of metric 3D localization, object identity consistency, explicit calibration, and multi-class mobile-object coverage within a common smart-space task. The 2026 release further adds paired appearance-transfer data and a hidden real-world target, allowing domain transfer to be evaluated without changing the task definition or annotation schema.

\begin{table}[t]
  \caption{Comparison of \paiss{} with representative public multi-camera benchmarks. OV/NOV indicates overlapping and non-overlapping camera views.}
  \vspace{-2mm}
  \label{tab:dataset_comparison_smartspaces}
  \centering
  \small
  \setlength{\tabcolsep}{3.5pt}
  \renewcommand{\arraystretch}{1.08}
  \resizebox{\linewidth}{!}{%
  \begin{tabular}{lrrrrllrrr}
    \toprule
    Dataset & Year & Frames & Cameras & IDs & Topology & Domain & 3D boxes & Depth & Resolution \\
    \midrule
    PETS2009~\citep{ferryman2009pets} & 2009 & 1.2K & 8 & 30 & OV & real outdoor & N/A & no & 768$\times$576 \\
    USC Campus~\citep{javed2008modeling} & 2010 & 135K & 3 & 146 & NOV & real outdoor & N/A & no & 852$\times$480 \\
    Passageway~\citep{berclaz2011multiple} & 2011 & 120K & 4 & 4 & OV & real outdoor & N/A & no & 320$\times$240 \\
    NLPR\_MCT~\citep{nlprmct2014} & 2015 & 355.5K & $\le$5 & $\le$235 & NOV & real indoor/outdoor & N/A & no & 320$\times$240 \\
    CamNet~\citep{zhang2015camnet} & 2015 & 360K & 8 & 50 & NOV & real indoor/outdoor & N/A & no & 640$\times$480 \\
    DukeMTMC~\citep{ristani2016performance} & 2016 & 2.45M & 8 & 2,834 & NOV & real outdoor & N/A & no & 1920$\times$1080 \\
    WILDTRACK~\citep{chavdarova2018wildtrack} & 2017 & 66.6K & 7 & N/A & both & real outdoor & N/A & no & 1920$\times$1080 \\
    MTA~\citep{kohl2020mta} & 2020 & 2.01M & 6 & 2,840 & both & synthetic & N/A & no & 1920$\times$1080 \\
    MultiviewX~\citep{hou2020multiviewx} & 2020 & 2.4K & 6 & 34 & OV & synthetic & N/A & no & 1920$\times$1080 \\
    MMPTRACK~\citep{han2023mmptrack} & 2021 & 2.98M & $\le$6 & $\le$140 & OV & real indoor & N/A & no & 640$\times$320 \\
    MTMMC~\citep{woo2024mtmmc} & 2024 & 3.05M & 16 & 3,669 & both & real indoor/outdoor & N/A & no & 1920$\times$1080 \\
    SCOUT~\citep{engilberge2025unified} & 2025 & 563.8K & 25 & 8,852 & both & real outdoor & N/A & no & 1920$\times$1080 \\
    \midrule
    \textbf{\paiss{} 2024} & 2024 & 22.87M & 953 & 2,481 & both & synthetic & 52.0M & no & 1920$\times$1080 \\
    \textbf{\paiss{} 2025} & 2025 & 4.54M & 504 & 363 & both & synthetic & 8.9M & yes & 1920$\times$1080 \\
    \textbf{\paiss{} 2026} & 2026 & 3.08M & 342 & 1,379 & both & synthetic + CT2.5 + real target & 11.1M & yes & 1920$\times$1080 \\
    \bottomrule
  \end{tabular}}
  \vspace{-7mm}
\end{table}

The comparison also clarifies why a new benchmark is useful despite the maturity of MTMC tracking. Real-world datasets such as DukeMTMC, WILDTRACK, MMPTRACK, MTMMC, and SCOUT provide valuable visual diversity and deployment realism, but their scale and annotation formats often reflect person tracking rather than physical AI perception. Synthetic resources such as MTA and MultiviewX make controlled annotation possible, but are smaller in camera-network scale and do not provide the combination of 3D boxes, depth, multi-class mobile objects, and Sim2Real stress testing required by current smart-space systems. \paiss{} occupies a complementary point in the design space: it does not replace real-world datasets, but supplies geometry-rich supervision and a standardized 3D evaluation protocol that would be difficult to obtain through manual labeling alone.

Three aspects are especially relevant for downstream method development. First, the dataset exposes methods to both overlapping and non-overlapping camera topologies, requiring systems to combine within-view tracking, cross-view association, and map-level reasoning rather than relying on a single cue. Second, the inclusion of non-human mobile objects changes the problem from person ReID to general object-level physical perception, where appearance features, size priors, motion patterns, and 3D geometry interact differently across classes. Third, the CT2.5 and real-target design makes visual domain shift an explicit evaluation variable. This separates the question of whether a model can exploit simulator geometry from the question of whether it can remain accurate under real camera appearance, lighting, and texture statistics.

\section{Dataset}
\label{sec:dataset}


\subsection{Scope}
\label{sec:scope}

\paiss{} targets indoor smart-space perception in environments such as warehouses, hospitals, and retail scenes, where multiple object classes move through a shared physical space and must be detected, localized, and consistently identified across a network of cameras. The full release contains 139 scenes, 1,799 cameras, and 282.5 hours of synchronized 1080p video, together with more than 72M 3D object annotations and 241M 2D bounding boxes spanning seven object classes (Table \ref{tab:dataset_scale}). 

Each scene is captured by a static, pre-calibrated camera network whose topology spans both overlapping and non-overlapping coverage, exposing blind spots, partial occlusions, and varying degrees of cross-camera overlap. The dataset is distributed as MP4/H.264 videos at 1080p and 30 FPS, together with per-camera calibration (intrinsics, extrinsics, and homographies to a shared world coordinate system), 3D bounding boxes, and object identities that are unique within each scene and class. All annotations are time-synchronized to the video frames and expressed in a single global world coordinate system per scene, so that 3D detection, tracking, and cross-camera association can be evaluated in metric units rather than in image space. The 2025 and 2026 components additionally provide per-camera depth maps, distributed as PNG sequences and packed HDF5 files. The Hugging Face dataset card documents the release structure, known issues, and CC-BY-4.0 license~\citep{physicalai_smartspaces_hf}.

\begin{table}[t]
  \caption{Scale of the \dataset{} release. The dataset provides synchronized 1080p video at 30 FPS, camera calibration, object identities, and dense 2D/3D supervision. \\{\tiny Object class abbreviations: P = Person, FL = Forklift, PT = PalletTruck, TP = Transporter, FG = FourierGR1T2, AD = AgilityDigit, and NC = NovaCarter.}}
  \vspace{-2mm}
  \label{tab:dataset_scale}
  \centering
  \small
  \setlength{\tabcolsep}{4pt}
  \resizebox{\linewidth}{!}{%
  \begin{tabular}{lrrrrrlrr}
    \toprule
    Component & Scenes & Hours & Cameras & Objects & Object classes & 3D labels & 2D boxes & Depth \\
    \midrule
    PAISS 2024 & 90 & 212.0 & 953 & 2,481 & P & 52.0M & 135.0M & no \\
    PAISS 2025 & 23 & 42.0 & 504 & 363 & P, FL, NC, TP, FG, AD & 8.9M & 73.0M & yes \\
    PAISS 2026 & 26 & 28.5 & 342 & 1,379 & P, FL, PT, NC, TP, FG, AD & 11.1M & 33.2M & yes \\
    \midrule
    Full release & 139 & 282.5 & 1,799 & 4,223 & 7 classes & 72.0M+ & 241.2M & partial \\
    \bottomrule
  \end{tabular}}
  \vspace{-6mm}
\end{table}

\subsection{Evaluation Claims and Assumptions}
\label{sec:evaluative_role}

The benchmark is designed to support the following evaluative claims:
\begin{itemize}
  \item \textbf{3D multi-camera perception:} models should jointly solve detection, localization, and association in a global world coordinate system, not only per-camera 2D tracking.
  \item \textbf{3D multi-class perception:} models should detect and track all 7 object classes: Person, Nova Carter, Transporter, FourierGR1T2, AgilityDigit, Forklift, and PalletTruck, rather than person-only.
  \item \textbf{Camera-network robustness:} models should handle both overlapping and non-overlapping views, occlusions, blind spots, and scene-scale movement.
  \item \textbf{Sim2Real generalization:} models should transfer from synthetic supervision and CT2.5 appearance-augmented RGB to the hidden real-world target, where depth is not assumed during inference.
\end{itemize}

These claims rely on several assumptions. First, simulator geometry, camera calibration, and automatic labels are treated as accurate enough to provide supervision for 3D localization and identity association. Second, the hidden real-world target is used to test deployment-oriented generalization, while the public synthetic and CT2.5 subsets provide inspectable training and validation data. Third, HOTA-style metrics are used because they jointly measure detection and association rather than optimizing one subproblem in isolation~\citep{luiten2021hota}.

The dataset is therefore best understood as an evaluation resource with explicit interpretive boundaries. Strong performance on the public synthetic validation split supports claims about geometry-aware learning under controlled annotation and calibration. Strong performance on CT2.5 scenes supports claims about robustness to appearance transfer under preserved layout and trajectories. Strong performance on the hidden real-world target supports a stronger, but still scoped, Sim2Real claim: the method can transfer to the real smart-space scenes represented by the benchmark without depth at inference time. None of these claims imply universal readiness for arbitrary surveillance deployments, outdoor traffic settings, or camera networks with substantially different optics and policies.

\subsection{Synthetic Generation and Curation}
\label{sec:curation}

\begin{figure}[t]
  \centering
  \includegraphics[width=\linewidth]{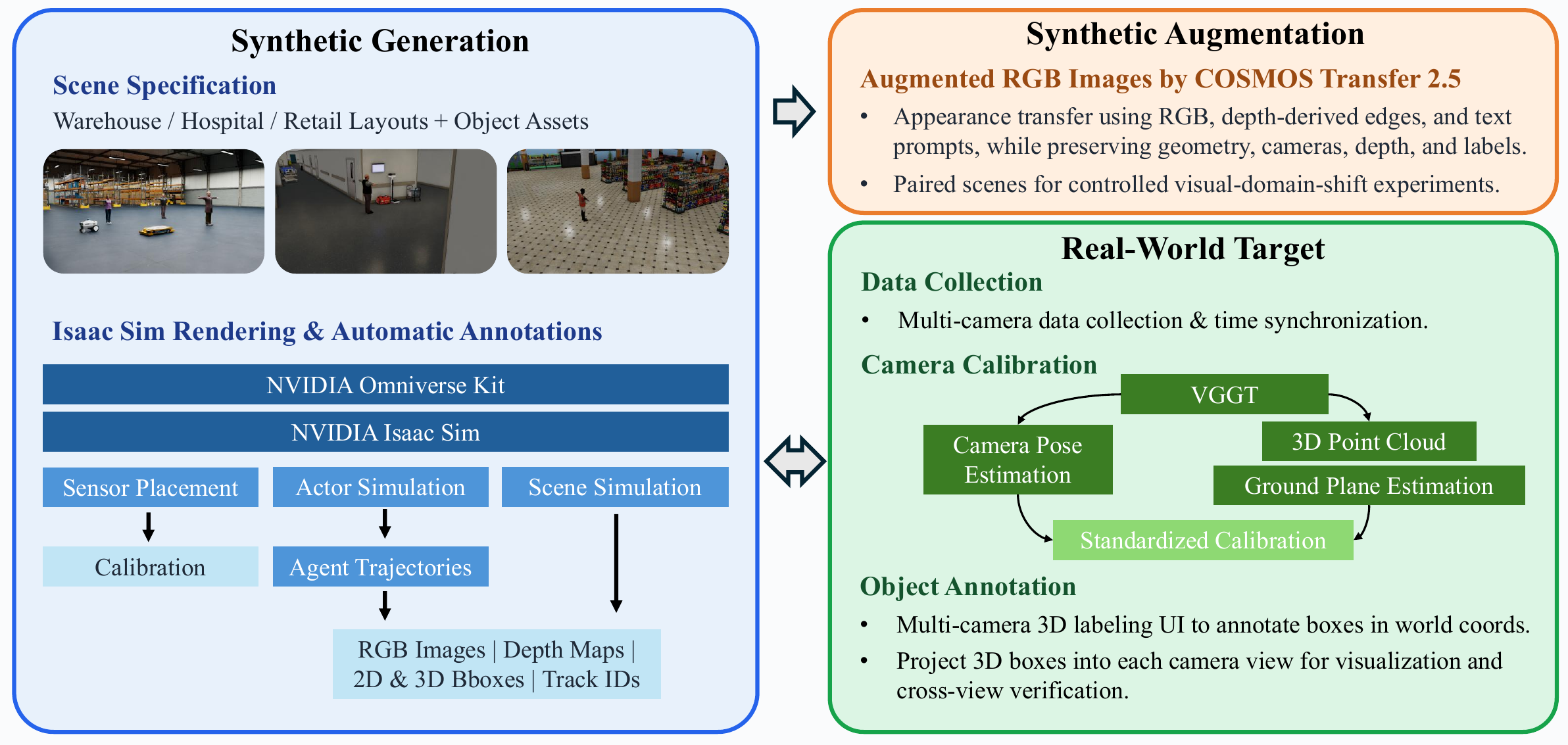}
  \vspace{-6mm}
  \caption{The \paiss{} pipeline. \textbf{Synthetic Generation}: Isaac Sim composes warehouse, hospital, and retail layouts with object assets and renders synchronized RGB/depth streams together with automatic 2D and 3D bounding boxes and track IDs. \textbf{Synthetic Augmentation}: Cosmos Transfer 2.5 takes the rendered RGB, depth-derived edges, and a text prompt to produce paired appearance-shifted RGB while preserving geometry, cameras, depth, and labels. \textbf{Real-World Target}: multi-camera streams are captured under time synchronization, calibrated through a VGGT-based pipeline that jointly estimates camera poses, 3D point clouds, and the ground plane into a standardized world coordinate system, and annotated via a multi-camera 3D labeling UI with cross-view box-projection verification.}
  \label{fig:generation_flow}
  \vspace{-6mm}
\end{figure}

The dataset is generated with NVIDIA Omniverse and Isaac Sim. We build on publicly released USD assets, including factory scenes together with person, robot, humanoid actors, etc., and assemble each of them on top of the Omniverse and Isaac Sim stack. The synthetic generation in Figure \ref{fig:generation_flow} illustrates the end-to-end pipeline. Four specific Isaac Sim tools are combined to configure agents, cameras, and synthetic data generation: IsaacSim Replicator Agent (IRA), Animated Robot Controller (IAR), RTX Sensor Placement (ISP), and RTX Sensor Calibration (ISC)~\citep{action_and_event_data_generation, nvidia_isaacsim_docs, tang20259th}. The full Synthetic Data Generation (SDG) pipeline that orchestrates these components into an end-to-end data generation workflow is documented separately~\citep{sdg_pipeline}

\paragraph{Scene and Actor Simulation.} Actor simulation is divided between IRA and IAR. IRA manipulates all human and humanoid actors, while IAR operates all wheeled platforms. Both tools expose multiple configuration modes: agents can wander randomly within a region, follow patrol trajectories along specified waypoints, or be activated by predefined trigger events. Scene simulation renders RGB images and depth maps, with the above assembled animations. This flexibility produces trajectories that deliberately exercise blind spots, partial occlusions, and changing camera overlap without requiring per-scene manual scripting.

\begin{figure}[t]
  \centering
  \includegraphics[width=\linewidth]{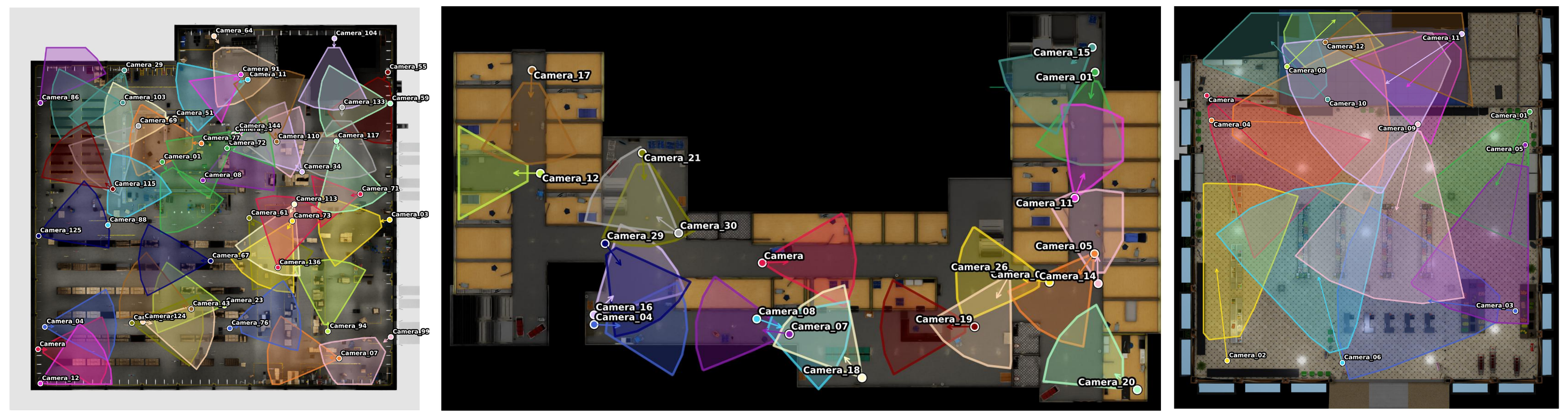}
  \vspace{-6mm}
  \caption{Top-down visualizations of PAISS camera networks across three scenes. Each colored cone depicts the field of view of one camera, with labels showing per-camera identifiers within the scene. To keep individual FOVs legible, the cameras shown here are downsampled from the full network in each scene; the released datasets contain a substantially denser deployment, with the full per-scene camera counts.}
  \label{fig:camera_network}
  \vspace{-4mm}
\end{figure}

\paragraph{Camera Calibration.} The camera network is constructed with ISP and ISC. ISP places cameras automatically based on the scene's floor plan, targeting either full ground coverage or a configurable coverage ratio; per-camera height, look-down angle, and coverage distance range are exposed as parameters. Once placement is verified, ISC generates the corresponding camera calibration, including intrinsics, extrinsics, projection matrices, and homographies to the shared global world coordinate system, so that all rendered streams are consistent in the same metric space. Figure~\ref{fig:camera_network} visualizes representative camera networks across three scenes.

\paragraph{Data Annotation and Curation.} Each rendered scene yields synchronized RGB and depth streams together with automatic per-frame annotations, including object identities, 2D visible bounding boxes, 3D bounding boxes, and calibration metadata, produced without any manual frame-level labeling (see the examples in Figure~\ref{fig:domain_comparison}). The raw renderings then pass through a curation stage before usage: we run automated sanity checks on frame synchronization, label-to-frame alignment, and identity continuity, and convert the raw outputs into the public dataset schema consumed by released models and evaluation tooling. 

\subsection{Synthetic Augmentation}
\label{sec:cosmos}

Although Isaac Sim provides geometry-rich synthetic data, models trained purely on it tend to overfit to simulator appearance, such as lighting, materials, and textures that systematically differ from real-world deployments. To broaden the appearance distribution seen during training and narrow the Sim2Real gap, we apply a synthetic augmentation stage on top of the Isaac Sim renderings using NVIDIA Cosmos Transfer 2.5 (CT2.5)~\citep{CT25doc,CT25repo}. The goal of this stage is to expose a wider range of visual conditions while keeping scene geometry, calibration, and ground-truth labels identical to the synthetic source, so that any difference in evaluation can be attributed to appearance rather than to changes in trajectories or annotations.

\begin{figure}[t]
  \centering
  \includegraphics[width=\linewidth]{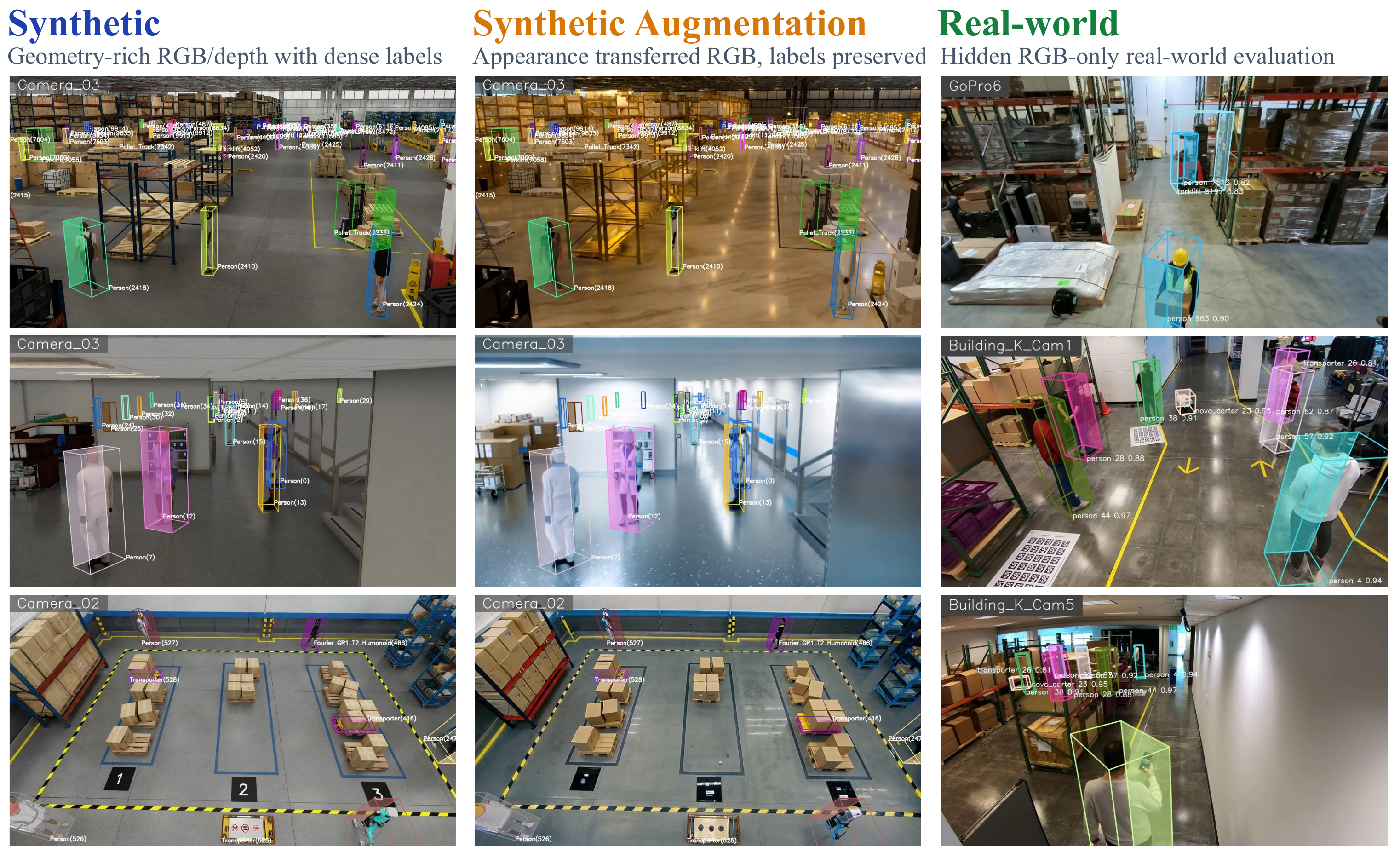}
  \vspace{-6mm}
  \caption{Visual domains in the benchmark. Isaac Sim data provides controllable synthetic scenes with dense labels; Cosmos Transfer changes appearance while preserving labels and calibration; Real-world scenes define the hidden RGB-only Sim2Real target.}
  \label{fig:domain_comparison}
  \vspace{-4mm}
\end{figure}

For each source scene, the CT2.5 pipeline takes three inputs: the generated RGB stream, an edge map derived from the corresponding depth map, and a text prompt describing the desired output style. CT2.5 then re-renders the scene under the prompted appearance while respecting the structural cues encoded by the RGB and edge inputs, producing an appearance-augmented RGB stream that remains pixel-aligned with the original calibration and labels. Representative outputs are shown in the Synthetic Augmentation of Figure~\ref{fig:domain_comparison}, where each augmented frame can be compared directly against its Isaac Sim source in the Synthetic.



\subsection{Real World Target}
\label{sec:realworld}

In addition to synthetic training data, our test split includes a real-world target domain captured from a deployed outside-in camera network. This split reflects the practical constraints of operating in unconstrained environments (e.g., changing lighting, clutter, and sensor variability) and is used to evaluate Sim2Real generalization.
The real-data pipeline comprises three stages (see Fig.~\ref{fig:camera_network}): (i) data collection, (ii) camera calibration, and (iii) object annotation.

\paragraph{Data collection.}
We capture two real-world warehouse deployments: one with 4 cameras and one with 7 cameras. All cameras are time-synchronized to enable consistent multi-view association and geometry.

\paragraph{Camera calibration.}
We estimate per-camera intrinsics and extrinsics automatically from synchronized multi-view video using VGGT~\cite{wang2025vggt}. VGGT reconstructs a dense point cloud and recovers each camera pose relative to a reference view; we then fit a dominant ground plane (e.g., via RANSAC) and normalize the coordinate frame so the origin lies on the ground and the $z$-axis aligns with gravity. The resulting parameters $(K_i, R_i, t_i)$ are used for multi-view projections, visibility estimation, and 3D box prediction.

\paragraph{Object annotation.}
We annotate 3D bounding boxes in the global world coordinate system using a dedicated UI. During labeling, each 3D box is projected into every camera view and overlaid on the synchronized videos, enabling cross-view visualization and verification of spatial consistency.

\section{Evaluation}
\label{sec:evaluation}

\subsection{Task Definition}
\label{sec:task_definition}

The main benchmark task is multi-camera 3D detection and tracking. Given synchronized videos from all cameras in a scene, a system must detect every relevant object and maintain a consistent identity as the object moves through time and across cameras. For the 2026 protocol, the object taxonomy is Person, NovaCarter, Transporter, FourierGR1T2, AgilityDigit, Forklift, and PalletTruck. The submission format is one detection per line:

\begin{center}
\texttt{scene\_id class\_id object\_id frame\_id x y z width length height yaw}
\end{center}

\noindent where coordinates are in the global world coordinate system, dimensions are in meters, and yaw is the object heading. Object identifiers must remain unique and consistent within each scene and class. The 2024 protocol evaluated 3D object locations, while the 2025 and 2026 protocols evaluate full 3D boxes using 3D IoU matching.

\subsection{3D HOTA Evaluation}
\label{sec:hota}

Evaluation uses 3D Higher Order Tracking Accuracy (HOTA), which jointly balances detection, association, and localization quality~\citep{luiten2021hota}. The original HOTA formulation is widely used for 2D multi-object tracking, where localization is measured by image-plane box overlap. PAISS instantiates the same detection--association principle in the global world coordinate system, making 3D HOTA a contribution of the benchmark protocol rather than only a choice of leaderboard metric. For a matching threshold $\alpha$, HOTA can be written as
\begin{equation}
\mathrm{HOTA}_{\alpha} = \sqrt{\mathrm{DetA}_{\alpha} \cdot \mathrm{AssA}_{\alpha}},
\label{eq:hota}
\end{equation}
where $\mathrm{DetA}_{\alpha}$ measures detection accuracy and $\mathrm{AssA}_{\alpha}$ measures association accuracy over matched trajectories. In PAISS, the similarity function used to form matches is defined over metric 3D locations or 3D boxes, so identity switches and localization errors are penalized in the physical scene rather than separately in each camera view.

In the 2024 protocol, HOTA was computed on 3D object locations after removing duplicate cross-camera observations for the same object and frame; Euclidean distances between predicted and ground-truth 3D locations were converted to similarity scores, with scores set to zero beyond 2~meters~\citep{wang20248th}. In the 2025 and 2026 protocols, matching instead uses 3D IoU between predicted and ground-truth boxes~\citep{tang20259th,aicity2026track1}. In all three protocols, HOTA is computed per class within each scene, averaged across classes, and then aggregated across scenes with weights proportional to the number of objects. The implementation builds on the TrackEval evaluation toolkit~\citep{trackeval}.

\subsection{Evaluation Server and Leaderboard Protocol}
\label{sec:server}

The AI City Challenge evaluation system provides a standardized, server-side implementation of the benchmark. Participants submit predictions to an online server, which checks the file format, runs the official metric implementation, and returns scores automatically~\citep{wang20248th,tang20259th}. During the active challenge, the leaderboard exposes only limited public feedback: top results are shown and each team can see its own current ranking. In 2024, in-challenge scores were computed on a random 50\% subset of the hidden test set to reduce leaderboard overfitting, with final rankings recomputed on the full hidden test set after the deadline; the 2025 challenge followed the same discipline, including submission limits and post-deadline full-test reporting~\citep{wang20248th}. The 2024 and 2025 evaluation servers remain open~\citep{aicity_dataset_access}, and the 2026 evaluation server will open with the 2026 challenge~\citep{aicity2026server}.

This server-side design fixes the data split, matching logic, class aggregation, and leaderboard score, avoiding small implementation differences that can change MTMC rankings. It also separates method development on public train/validation data from generalization measurement on hidden test data, and reports raw HOTA on the public leaderboard while applying the 10\% online-tracker bonus only when determining final challenge awards for methods whose paper and code demonstrate no use of future frames~\citep{aicity2026track1}.

The server further encodes a data-use policy: teams competing for awards may not manually label the test set or use private data on the public leaderboard~\citep{wang20248th}, and top-ranked teams are expected to release code or sufficient implementation detail so that leaderboard results can be audited and reused~\citep{tang20259th}. Together with the HOTA decomposition into detection, localization, and association terms, this protocol allows PAISS to compare qualitatively different pipelines, detector-first, geometry-first, ReID-first, and hybrid online/offline systems, without collapsing the task to a single per-camera detection score.

\subsection{Reference Results from AI City Challenge Track 1}
\label{sec:results}

The AI City Challenge provides an empirical record of how systems perform as the benchmark has shifted from person-only 3D location tracking to multi-class 3D box tracking. Tables~\ref{tab:track1_2024} and~\ref{tab:track1_2025} reproduce the Track 1 leaderboards reported in the 2024 and 2025 challenge papers. These results serve as reference points for future methods and illustrate that both offline global association and online tracking pipelines remain important.

\begin{table}[t]
  \caption{AI City Challenge 2024 Track 1 leaderboard for MTMC people tracking. Scores are reported as HOTA in the challenge paper~\citep{wang20248th}.}
  \label{tab:track1_2024}
  \centering
  \small
  \resizebox{\linewidth}{!}{%
  \begin{tabular}{lrlcc}
    \toprule
    Team & Team ID & Representative method paper & HOTA & Online \\
    \midrule
    Met\_BEV & 331 & MCBLT~\citep{wang2025mcblt} & 81.2182 & no \\
    Yachiyo & 221 & overlap suppression clustering~\citep{yoshida2024overlap} & 71.9446 & no \\
    SJTU-Lenovo & 79 & geometric consistency and state-aware ReID~\citep{xie2024robust} & 67.7258 & yes \\
    Nota & 40 & cluster self-refinement~\citep{kim2024cluster} & 62.1666 & yes \\
    Fraunhofer IOSB & 142 & OCMCTrack~\citep{specker2024ocmctrack} & 60.8792 & yes \\
    UW-ETRI & 8 & online long-sequence tracking~\citep{yang2024online} & 57.1445 & yes \\
    Asilla & 162 & memory-efficient appearance features and trajectory validation~\citep{vi2024efficient} & 55.8806 & yes \\
    ARV & 50 & spatial-temporal anchors and hierarchical clustering~\citep{suttichaya2024online} & 51.0556 & yes \\
    \bottomrule
  \end{tabular}}
  \vspace{-6mm}
\end{table}

The 2024 leaderboard reflects a mature recipe for person-centric MTMC tracking: high-quality 2D detection, ReID embeddings, pose- or footpoint-based projection into a shared coordinate system, and tracklet association across cameras. Most teams used YOLO-family detectors and modern ReID backbones, and nearly all systems used pose estimation to improve 3D footpoint localization~\citep{wang20248th}. The top-1 method MCBLT~\citep{wang2025mcblt}, however, introduces a BEV-based framework, which first aggregates multi-view images with necessary camera calibration parameters to obtain 3D object detections, then proposes hierarchical GNNs to track these 3D detections. Unlike other existing methods, MCBLT has impressive generalizability across different scenes and diverse camera settings, with exceptional capability for long-term association handling. Another leading method Yachiyo uses overlap suppression clustering and selective ReID over representative, recognizable tracklet images~\citep{yoshida2024overlap}. Its advantage shows that access to future frames and global association can still substantially improve identity consistency in long, multi-camera sequences.

The online systems are nevertheless important for physical AI deployment. SJTU-Lenovo combined geometric consistency with state-aware ReID correction~\citep{xie2024robust}; Nota refined anchors through appearance clustering and duplicate-overlap correction~\citep{kim2024cluster}; Fraunhofer IOSB used a corrective matching cascade~\citep{specker2024ocmctrack}.
Taken together, these methods indicate that \paiss{} stresses the trade-off between global association quality and online operation: systems must maintain compact global state, correct local tracking mistakes, and avoid duplicate identities when multiple cameras observe the same person.

\begin{table}[t]
  \caption{AI City Challenge 2025 Track 1 leaderboard for multi-class 3D multi-camera tracking. Scores are reported as raw HOTA in the challenge paper~\citep{tang20259th}.}
  \label{tab:track1_2025}
  \centering
  \small
  \resizebox{\linewidth}{!}{%
  \begin{tabular}{lrlccc}
    \toprule
    Team & Team ID & Representative method paper & HOTA & Online & Input Modalities \\
    \midrule
    ZV & 65 & 3D point clouds and ReID~\citep{lee2025multi} & 69.9118 & no & RGB and Depth \\
    Metropolis-Sparse4D & 220 & A unified 3D detection and tracking network~\citep{wang2026unified} & 67.3143 & yes & RGB \\
    SKKU-AutoLab & 15 & DepthTrack~\citep{tran2025depthtrack} & 63.1396 & no & RGB and Depth \\
    TeamQDT & 133 & Robust 2D tracking with late depth aggregation~\citep{vuminh2025online} & 31.6267 & yes & RGB and Depth \\
    UTE AI Lab & 116 & VGCRTrack~\citep{phan2025vgcrtrack} & 28.0345 & yes & RGB and Depth \\
    \bottomrule
  \end{tabular}}
  \vspace{-6mm}
\end{table}

The 2025 results demonstrate how the benchmark becomes more demanding when the target changes from person locations to multi-class 3D boxes. The winning ZV system fused depth maps into a unified point cloud, used transformer-based 3D detection, and addressed rare classes with 3D shape embeddings before applying hybrid online/offline association~\citep{lee2025multi,tang20259th}. Metropolis-Sparse4D~\citep{wang2026unified} is the only online method with RGB-image-only input, which impressively achieves a very high HOTA score compared with other methods consume both RGB images and depthmaps.  SKKU-AutoLab's DepthTrack connected BEV-space tracklets to clustered point clouds through Tracklet-Cluster Mapping, while adding ViT-based ReID, pose-guided yaw estimation, and visual enhancement for robustness~\citep{tran2025depthtrack}. These two leading offline approaches show that depth, calibration, and sparse 3D reasoning can be exploited aggressively when the full sequence is available.
TeamQDT extended robust 2D MTMC tracking into 3D by preserving 2D ID consistency and aggregating depth later in the pipeline~\citep{vuminh2025online}. UTE AI Lab's VGCRTrack used view-aware geometric center refinement and trajectory-level affinity metrics, including 3D IoU, to improve localization and association under occlusion~\citep{phan2025vgcrtrack}. 

\section{Conclusion}
\label{sec:conclusion}

\dataset{} provides a large-scale benchmark for evaluating multi-camera 3D perception in physical AI smart spaces. By combining synthetic supervision, dense 2D/3D labels, calibration, depth, Cosmos Transfer appearance augmentation, and a hidden real-world Sim2Real evaluation, the dataset supports a sharper evaluation question than scale alone: can a system maintain object identity and 3D localization across complex camera networks when moving from controlled generated data to deployment-like visual conditions? The 2024 and 2025 leaderboards establish reference performance, while the 2026 benchmark broadens the task toward RGB-only Sim2Real generalization for multi-class indoor perception.

The benchmark is bounded in scope and intended for careful research use. The data covers indoor smart spaces captured by static, pre-calibrated 1080p cameras, and is dominated by warehouse environments; appearance is rendered by Cosmos Transfer 2.5, which is single-view and inherits its own training-data biases. The synthetic avatars are drawn from a finite Omniverse asset library that is not balanced across body type, skin tone, age, or mobility, so the dataset is not suitable for fairness or demographic studies, and is not authorized for biometric, facial recognition, or surveillance use against real individuals. The public splits contain no real-person imagery; the 2026 real-world Sim2Real target was captured under signed individual consent and is held out behind the AI City Challenge evaluation server, with manual test labeling prohibited and the target excluded from training. Strong leaderboard performance should be read as evidence of geometry-aware multi-camera tracking under the benchmark's conditions, not as deployment readiness. Real-site use requires privacy review, access controls, and validation in the target environment. Within these boundaries, the dataset offers a reproducible, geometry-aware bridge between controllable synthetic generation, appearance transfer, and real-world evaluation.

\newpage
\bibliographystyle{plainnat}
\bibliography{main}


\end{document}